\documentclass[11pt,a4paper]{article}

\usepackage[margin=1in]{geometry}
\usepackage{graphicx}
\usepackage{booktabs}
\usepackage{longtable}
\usepackage{array}
\usepackage{amsmath}
\usepackage{amssymb}
\usepackage[square,numbers]{natbib}  
\usepackage{hyperref}
\usepackage{multirow}
\usepackage{titling}      
\usepackage{xcolor} 

\usepackage{longtable}
\usepackage{siunitx}
\usepackage{array}
\usepackage{booktabs}
\usepackage{float}
\usepackage{placeins}
\usepackage[most]{tcolorbox}
\usepackage{caption}

\usepackage{titlesec}
\titlespacing*{\section}{0pt}{1.5em}{0.8em}
\titlespacing*{\subsection}{0pt}{1.2em}{0.6em}

\usepackage{caption}
\usepackage{fancyhdr}
\makeatletter
\def\@maketitle{%
	\newpage
	\begin{center}%
		\let\footnote\thanks
		{\LARGE\bfseries \@title \par}%
		\vspace{1.7em}
		{\large \@author \par}%
		\vspace{1.5em}
	\end{center}%
	\par
}
\makeatother

\makeatletter
\renewcommand\include[1]{\@input{#1}}
\makeatother

\definecolor{linkblue}{RGB}{0,71,171}

\hypersetup{
	colorlinks=true,
	linkcolor=linkblue,
	citecolor=linkblue,
	urlcolor=linkblue,
	filecolor=linkblue,
}

\title{\Large\bfseries PRICE: A Systematic Study of LLM Adaptation Choices for Bitcoin Price Forecasting}

\author{
	\begin{minipage}[t]{0.45\textwidth}
		\centering
		\textbf{Maryam Fakhari}\\
		{\small Department of Electrical and Computer Engineering}\\
		{\small Isfahan University of Technology}\\
		{\small Isfahan, Iran}\\
		\texttt{\color{linkblue}m.fakhari@ec.iut.ac.ir}
	\end{minipage}
	\hfill
	\begin{minipage}[t]{0.45\textwidth}
		\centering
		\textbf{Mehran Safayani}\\
		{\small Department of Electrical and Computer Engineering}\\
		{\small Isfahan University of Technology}\\
		{\small Isfahan, Iran}\\
		\texttt{\color{linkblue}safayani@iut.ac.ir}
	\end{minipage}
}
\date{}

\begin{document}
	
	\let\WriteBookmarks\relax
	\def\floatpagepagefraction{1}
	\def\textpagefraction{.001}
	
	\newtcolorbox{promptbox}{
		width=0.95\linewidth,
		colback=gray!6,
		colframe=black,
		boxrule=0.6pt,
		arc=2mm,
		fontupper=\ttfamily\small,
		left=8pt,
		right=8pt,
		top=8pt,
		bottom=8pt
	}
	
	\maketitle

	\begin{abstract}
		Cryptocurrency markets exhibit extreme volatility and non-stationary dynamics that challenge conventional forecasting methods. Although Large Language Models (LLMs) have shown promise for time series forecasting, the combined effects of adaptation choices remain largely unexplored in financial environments. 
		This study introduces PRICE, a structured approach for adapting LLMs to short-term Bitcoin price forecasting. 
		Built on a 4-bit quantized LLaMA-3 8B model, PRICE investigates how parameter-efficient fine-tuning, numerical representation, prompting strategy, inference strategy, and decoding jointly influence forecasting performance.
		PRICE integrates \textbf{P}arameter-efficient fine-tuning with Low-Rank Adaptation (LoRA), \textbf{R}ecursive multi-step inference, \textbf{I}nteger-rounded numerical representation, \textbf{C}ontext-Task-Format (CTF) prompting, and \textbf{E}xact zero-temperature decoding. Ablation studies show that each adaptation choice contributes measurably to forecasting accuracy and prediction reliability. LoRA enables efficient training and inference on limited hardware, recursive inference improves predictive accuracy, integer-rounded price values consistently reduce forecasting errors, CTF prompting outperforms Chain-of-Thought, Implicit Chain-of-Thought (iCoT), and few-shot prompting even after fine-tuning, and exact zero-temperature decoding improves output stability during recursive forecasting.
		Comparative evaluation against eight transformer-based and time-series foundation models demonstrates that PRICE achieves the lowest forecasting errors on both validation and test sets while maintaining robust performance across evaluation periods where several baselines exhibit substantial degradation. Notably, despite being based on a model primarily pretrained on text rather than time-series data, PRICE achieves competitive or superior performance relative to specialized time-series foundation models. These findings demonstrate that adaptation choices, often regarded as secondary implementation details, critically determine forecasting accuracy and robustness when adapting large language models to numerical time series forecasting tasks.
	\end{abstract}


	\section{Introduction}
	
	Accurate time series forecasting plays a critical role in data-driven decision-making across finance, economics, energy systems, and industrial applications \cite{mahalakshmi2016survey}. In financial markets, reliable forecasting models enhance risk management, trading strategies, and capital allocation. However, forecasting performance deteriorates significantly in highly volatile environments where price dynamics are nonlinear, non-stationary, and strongly influenced by external factors.
	
	Cryptocurrency markets represent one of the most challenging instances of such environments. Compared to traditional financial assets, cryptocurrencies exhibit substantially higher volatility, abrupt regime shifts, and sensitivity to speculative behavior and social sentiment \cite{john2024cryptocurrency}. These characteristics amplify uncertainty and limit the effectiveness of conventional forecasting techniques, necessitating more adaptive and robust modeling strategies tailored to extreme market dynamics.
	
	Traditional statistical methods and machine learning models frequently struggle to capture the complex nonlinear dependencies and evolving temporal patterns present in cryptocurrency time series \cite{mahalakshmi2016survey}. More recently, Transformer-based architectures have demonstrated strong capabilities for modeling long-range dependencies through self-attention mechanisms. Nevertheless, their performance in highly volatile financial environments remains inconsistent, with challenges including sensitivity to noise, error accumulation in multi-step forecasting, and limited robustness across different market regimes \cite{yu2023temporal}.
	
	Building on these developments, foundation models have introduced a new paradigm in machine learning by leveraging large-scale pre-training to learn transferable representations across tasks, rather than training task-specific models from scratch. While this paradigm has achieved remarkable success in natural language processing and computer vision, its direct application to time series forecasting remains non-trivial, since time series lack explicit semantic structure, exhibit strong domain heterogeneity, and often provide limited labeled samples; these challenges are further intensified in financial time series by structural breaks, heavy-tailed distributions, and extreme volatility \cite{miller2024survey}.
	
	Recent advances in Large Language Models (LLMs) have opened new opportunities for time series forecasting. Owing to their strong sequence modeling, reasoning, and generalization capabilities, LLMs have increasingly been adapted to forecasting tasks beyond natural language processing. In financial applications, existing LLM-based approaches generally follow three directions: zero-shot or few-shot adaptation, domain-specific fine-tuning, and specialized pre-training on financial corpora \cite{yu2023temporal, li2023large}. 
	
	Existing studies have primarily investigated individual aspects of LLM adaptation in isolation, such as numerical tokenization strategies \cite{gruver2023llmtime}, parameter-efficient fine-tuning techniques for financial forecasting \cite{usmanfine}, or prompt-based reformulation of forecasting tasks \cite{xue2022promptcast}. While these studies have demonstrated the potential of LLMs for time series analysis, they provide limited insight into how multiple adaptation choices interact within a unified forecasting framework. In particular, the combined effects of numerical input representation, prompt design, learning paradigm, decoding strategy, and model adaptation on forecasting performance remain largely unexplored in cryptocurrency markets. Consequently, it is still unclear which factors contribute most significantly to forecasting accuracy and robustness when LLMs are applied to highly volatile and non-stationary financial data.
	
	To address this gap, this study proposes \textbf{PRICE}, a systematic framework for adapting large language models to short-term Bitcoin price forecasting. Built on a 4-bit quantized LLaMA-3 8B model and parameter-efficient fine-tuned using Low-Rank Adaptation (LoRA), PRICE investigates five methodological components: \textbf{P}arameter-efficient fine-tuning, \textbf{R}ecursive multi-step inference, \textbf{I}nteger-rounded numerical representation, \textbf{C}ontext-Task-Format (CTF) prompting, and \textbf{E}xact zero-temperature decoding. Specifically, the framework integrates these five components into a unified forecasting pipeline to examine how different adaptation choices influence forecasting performance. Rather than focusing solely on overall model performance, this study conducts controlled ablation experiments to isolate the contribution of each adaptation choice and evaluate its effect on forecasting accuracy and prediction reliability. This systematic analysis provides empirical evidence on how individual adaptation decisions affect the accuracy and robustness of LLMs when applied to highly volatile and non-stationary time series.
	
	The main contributions of this study are summarized as follows:
	\begin{itemize}
		
		\item We propose PRICE, a unified framework for adapting large language models to short-term Bitcoin price forecasting. PRICE integrates parameter-efficient fine-tuning, recursive multi-step inference, integer-rounded numerical representation, Context-Task-Format (CTF) prompting, and exact zero-temperature decoding within a single forecasting pipeline.
		
		\item We develop a controlled ablation framework to systematically quantify the contribution of individual adaptation choices, including learning paradigm, forecasting strategy, numerical representation, prompt design, and decoding strategy, under consistent experimental conditions.
		
		\item We demonstrate that parameter-efficient fine-tuning and recursive multi-step inference substantially improve LLM-based Bitcoin forecasting. Fine-tuning enhances predictive accuracy, while recursive inference improves both forecasting performance and output reliability compared with direct multi-step generation.
		
		\item We reveal the importance of numerical representation for LLM-based financial forecasting. Integer-rounded representations consistently outperform raw, standard-normalized, and instance-normalized inputs, suggesting that preserving the original price scale while reducing unnecessary precision provides a more effective input representation.
		
		\item We investigate the role of prompt formulation and decoding strategy in adapted LLMs. The results show that CTF prompting remains effective after fine-tuning, outperforming Chain-of-Thought (CoT), Implicit Chain-of-Thought (iCoT), and few-shot prompting.
		
		\item We validate the effectiveness and robustness of PRICE through comparisons with eight transformer-based and time-series foundation models. PRICE achieves the lowest forecasting errors among the evaluated methods on both validation and test sets and maintains reliable performance across different market periods.
		
	\end{itemize}
	
	The remainder of this paper is organized as follows. Section 2 reviews related work in statistical and neural forecasting, transformer-based and time-series foundation models, and large language models for time series forecasting. Section 3 presents the PRICE methodology and describes its five adaptation components, including parameter-efficient fine-tuning, recursive multi-step inference, integer-rounded numerical representation, CTF prompting, and exact zero-temperature decoding. Section 4 details the experimental setup, including dataset construction and preprocessing, model configuration, implementation details, and evaluation metrics. Section 5 presents the ablation studies and comparative evaluation against transformer-based and time-series foundation models on the validation and test sets. Finally, Section 6 concludes the paper and discusses its limitations and directions for future research.
	
	
	\section{Related Works}
	
	Existing approaches to financial time series forecasting can generally be categorized into three main groups: traditional statistical methods, machine learning–based models, and deep learning approaches. The following subsections review these categories and highlight their respective strengths and limitations.

	\subsection{Statistical and Neural Forecasting Methods}
	
	Early approaches to financial time series forecasting primarily relied on statistical modeling techniques such as Autoregressive (AR) and Autoregressive Integrated Moving Average (ARIMA) models~\cite{box2015time}, which were widely adopted due to their interpretability and analytical simplicity. Exponential smoothing methods have also been applied for short-term forecasting because of their computational efficiency~\cite{liu2021forecast}. However, these approaches generally assume linear relationships, stationarity, and stable distributions, assumptions that are often violated in highly volatile and non-stationary financial markets, limiting their predictive capability and adaptability~\cite{kontopoulou2023review}.
	
	Machine learning approaches significantly extended forecasting capabilities by relaxing these assumptions. Early data-driven models such as support vector machines~\cite{vapnik1995nature} and multi-layer perceptrons~\cite{hornik1989multilayer} demonstrated greater flexibility in modeling complex temporal patterns compared to classical statistical approaches. With the emergence of deep learning, long short-term memory (LSTM) networks~\cite{hochreiter1997long} tailored for sequence modeling and temporal convolutional networks (TCNs)~\cite{bai2018empirical} designed for longer temporal dependencies achieved strong forecasting performance by capturing nonlinear and long-range temporal relationships. Nevertheless, these approaches generally remain task-specific and often exhibit limited generalizability across diverse time series forecasting scenarios~\cite{kontopoulou2023review,jin2024timellm}.
	
	\subsection{Transformer-Based Architectures and Foundation Models}
	
	Transformer-based architectures~\cite{vaswani2017attention} have gained significant attention in financial and cryptocurrency time series forecasting due to their ability to capture long-range dependencies through self-attention mechanisms. Compared to recurrent and convolutional models, transformers offer improved scalability and flexibility, making them well-suited for highly volatile and nonlinear market environments. Several transformer variants have been proposed to better accommodate the structural characteristics of time series data. Pyraformer~\cite{liu2022pyraformer} improves efficiency for long-horizon forecasting, Autoformer~\cite{wu2021autoformer} incorporates trend--seasonal decomposition to enhance performance on non-stationary data, and PatchTST~\cite{nie2022time} employs patch-based representations to capture multi-scale temporal dependencies. Other models such as TimesNet~\cite{wu2022timesnet} and Crossformer~\cite{zhang2023crossformer} further extend transformer architectures through alternative temporal representations and cross-variable dependency modeling. Nevertheless, challenges related to robustness under extreme market volatility and generalization across market regimes remain~\cite{zeng2023transformers}.
	
	Building on these developments, time series foundation models aim to leverage large-scale pre-training across heterogeneous datasets to enable general-purpose and zero-shot forecasting~\cite{miller2024survey}. Lag-LLaMA~\cite{rasul2023lag} adapts large language model architectures for probabilistic forecasting via extensive pre-training on diverse time series corpora. Chronos~\cite{ansari2024chronos} adopts a tokenization-based formulation that converts continuous signals into discrete representations, enabling compatibility with language model architectures and supporting zero-shot inference. However, these models still face limitations stemming from strong domain heterogeneity, limited availability of high-quality labeled time series, and the lack of universally consistent temporal representations~\cite{miller2024survey}. 
	
	\subsection{Large Language Models for Time Series Forecasting}
	
	The success of large language models (LLMs) in natural language processing has motivated growing interest in their application to time series analysis, where continuous numerical signals differ fundamentally from textual data. Existing approaches span diverse paradigms for adapting LLMs to temporal modeling~\cite{zhang2024llmtimeseries}. Among these, our work is most closely related to direct prompting and fine-tuning approaches.
	
	Direct prompting reformulates numerical sequences as text and queries pre-trained LLMs without parameter updates~\cite{zhang2024llmtimeseries}. Early work such as PromptCast~\cite{xue2022promptcast} framed prediction as a sentence-to-sentence generation task. LLMTime~\cite{gruver2023llmtime} improved numerical tokenization through digit-level representations, demonstrating competitive zero-shot performance with GPT-3 and LLaMA-2. In financial prediction, prior studies have explored zero-shot prompting of ChatGPT using market information and explainable reasoning mechanisms for temporal modeling~\cite{xie2023wallstreet,yu2023temporaldata}.
	
	A complementary research direction aligns time series representations with pre-trained LLMs or adapts LLM backbones for temporal prediction. Representative approaches include GPT4TS~\cite{zhou2023onefitsall}, Time-LLM~\cite{jin2024timellm}, and AutoTimes~\cite{liu2024autotimes}, which introduce lightweight adaptation mechanisms while preserving most pre-trained parameters. More recently, parameter-efficient fine-tuning methods such as LoRA~\cite{hu2022lora} have gained attention. LLIAM~\cite{german2025transfer} reformulates time series as textual prompts for LoRA-tuned LLaMA, while Time-LlaMA~\cite{zhang2025time} introduces dynamic low-rank adaptation for temporal prediction. In financial applications, Usman et al.~\cite{usmanfine} apply LoRA fine-tuning to LLaMA-based models for stock price prediction.
	
	Our approach is most closely related to LoRA-based LLM forecasting methods that employ textual representations of numerical sequences. However, prior studies primarily focus on general benchmarks or traditional financial assets, while cryptocurrency forecasting and the systematic evaluation of prompting, representation, and inference strategies remain comparatively underexplored.
	
	\section{Methodology}
	
	\begin{figure*}[t]
		\centering
		\includegraphics[height=10.5cm,width=17cm]{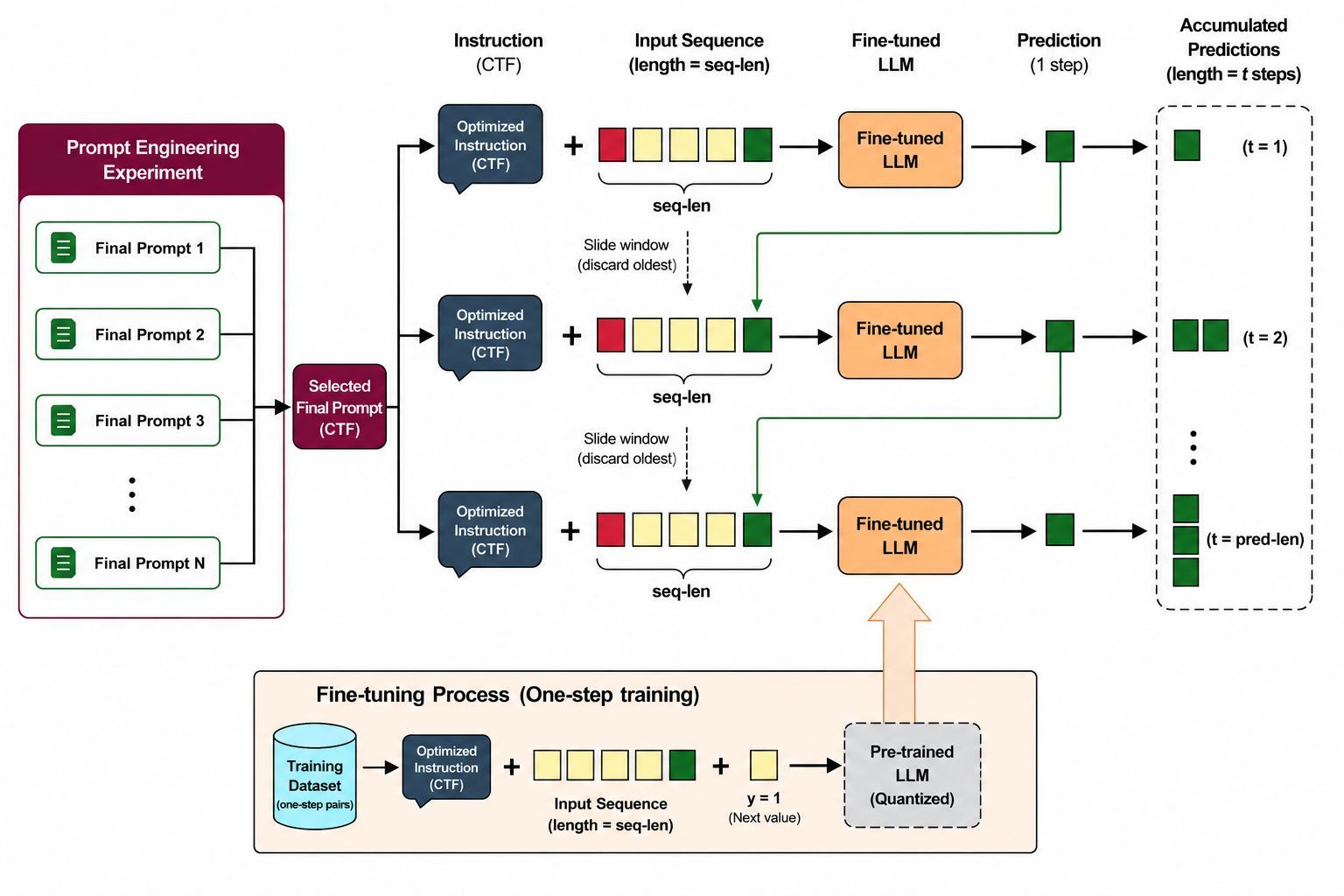}
		\caption{Overall workflow of the PRICE framework for adapting a quantized LLaMA-3 8B model to short-term Bitcoin price forecasting.}
		\label{proposed_method}
	\end{figure*}
	
	This section presents PRICE, a systematic framework for adapting large language models to short-term Bitcoin price forecasting. PRICE investigates five complementary adaptation components: Parameter-efficient fine-tuning, Recursive multi-step inference, Integer-rounded numerical representation, Context-Task-Format (CTF) prompting, and Exact zero-temperature decoding. These components are integrated into a unified forecasting pipeline based on a 4-bit quantized LLaMA-3 8B model to examine how different adaptation choices influence forecasting accuracy and prediction reliability. The overall workflow of PRICE, including prompt selection, one-step fine-tuning, recursive multi-step inference, and the generation of the final forecasting horizon, is illustrated in Fig.~\ref{proposed_method}. The methodological details of each component are presented in the following subsections.
	
	\subsection{P: Parameter-Efficient Fine-Tuning with LoRA}
	
	During preprocessing, the training dataset is organized using a fixed input sequence length (\texttt{seq\_len}) and prediction length (\texttt{pred\_len}). During the training phase, the prediction length is set to one (\texttt{pred\_len}=1), such that each training instance consists of a historical price sequence and its corresponding next-step target:
	
	\begin{align}
		X &= [x_1, x_2, \dots, x_{\text{seq\_len}}], \\
		Y &= [y_1].
	\end{align}
	
	The language model is adapted using Low-Rank Adaptation (LoRA)~\cite{hu2022lora}, a parameter-efficient fine-tuning technique that introduces trainable low-rank matrices into selected transformer layers while keeping the majority of the pre-trained parameters frozen. This strategy substantially reduces computational and memory requirements while enabling effective adaptation of the LLM to Bitcoin price dynamics.
	
	Rather than directly optimizing the model for multi-step forecasting, PRICE adopts a one-step-ahead training objective. This formulation is consistent with the autoregressive generation mechanism of language models, where each prediction is conditioned on previously observed context.
	
	\subsection{R: Recursive Multi-Step Inference}
	
	At inference time, PRICE employs a recursive autoregressive forecasting strategy to generate multiple future price values. The input sequence length remains identical to the training phase, while the forecasting horizon (\texttt{pred\_len}) can be adjusted according to the prediction task.
	
	The model first receives a fixed-length historical sequence and generates a one-step-ahead prediction. The generated value is then appended to the input sequence, while the oldest observation is removed to maintain a fixed context length. This updated sequence is recursively provided to the model until the desired forecasting horizon is reached.
	
	Formally, given an initial input sequence:
	
	\[
	X^{(0)} =
	[x_1, x_2, \dots, x_{\text{seq\_len}}],
	\]
	
	the prediction at step \(t\) is generated as:
	
	\[
	\hat{y}_t = f(X^{(t-1)}),
	\]
	
	where \(f(\cdot)\) denotes the fine-tuned language model. After each prediction, the input sequence is updated as:
	
	\[
	X^{(t)}
	=
	[x_{t+1}, \dots, x_{\text{seq\_len}}, \hat{y}_1, \dots, \hat{y}_t].
	\]
	
	This recursive procedure continues until \texttt{pred\_len} future values are generated. By extending the one-step training objective through autoregressive inference, PRICE enables flexible multi-step forecasting while remaining consistent with the sequential prediction mechanism of language models.
	
	\subsection{I: Integer-Rounded Numerical Representation}
	
	A central component of PRICE is the numerical representation of Bitcoin prices provided to the language model. Unlike conventional time series models that directly operate on continuous numerical values, LLMs process input sequences through tokenization mechanisms originally developed for natural language. Consequently, the representation and precision of numerical values can influence how effectively an LLM processes and forecasts financial time series.
	
	To construct the forecasting samples, a sliding-window strategy is employed to transform the time series into input--target pairs. For each sample, a subsequence of length \texttt{seq\_len} is used as the historical input, while the subsequent \texttt{pred\_len} observations constitute the forecasting targets. Formally, each instance is defined as
	
	\begin{align}
		\text{Input: } & [p_i, p_{i+1}, \dots, p_{i+\text{seq\_len}-1}], \\
		\text{Target: } & [p_{i+\text{seq\_len}}, \dots, 
		p_{i+\text{seq\_len}+\text{pred\_len}-1}].
	\end{align}
	
	The sliding window advances with a stride of one, allowing the available historical observations to be utilized while preserving the temporal ordering of the data.
	
	Raw Bitcoin prices may contain long decimal expansions, such as \texttt{69340.0606666667}. When converted into text, such values can be fragmented into multiple tokens by the LLM tokenizer, potentially introducing unnecessary representational complexity. PRICE therefore represents Bitcoin prices using integer-rounded values, preserving the original price scale while removing unnecessary decimal precision.
	
	Specifically, each price value is rounded to the nearest integer before being converted into textual form. For example, with \texttt{seq\_len}=96 hourly closing prices, an input sequence is represented as:
	
	\begin{quote}
		\small
		\texttt{69340, 69570, 69695, 69519, 69543, 69505, \dots, 64304, 64438}
	\end{quote}
	
	The resulting integer sequence is directly incorporated into the forecasting prompt as the numerical context provided to the language model. The effectiveness of this representation relative to raw, standard-normalized, and instance-normalized alternatives is evaluated through the ablation experiments presented in Section~\ref{Effect-of-Input-Representation}.

	\subsection{C: Context-Task-Format Prompting}
	\label{Forecasting-Instruction}
	
	\begin{figure*}[t]
		\centering
		\includegraphics[height=6.75cm,width=\textwidth]{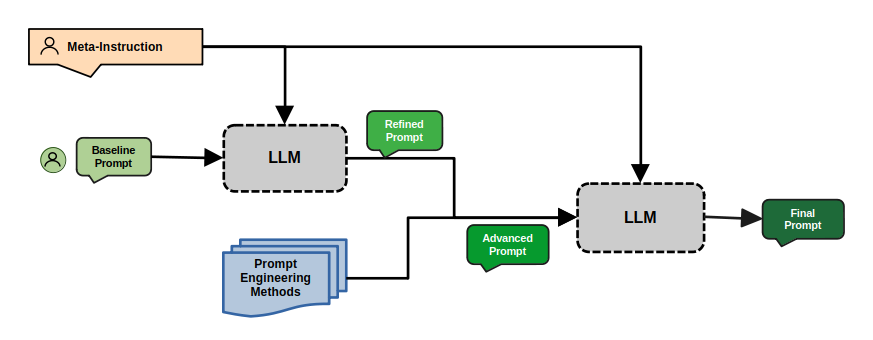}
		\caption{Stages of instruction preparation for Context-Task-Format (CTF) prompting.}
		\label{prompt-engineering}
	\end{figure*}
	
	\begin{figure}[ht]
		\centering
		\begin{promptbox}
			Using the following hourly Bitcoin closing prices, predict the next closing price based on comprehensive technical analysis.\\
			Return only one predicted number --- no explanation, no text.\\
			-- input series --
		\end{promptbox}
		\caption{Final Context-Task-Format (CTF) forecasting prompt used in PRICE.}
		\label{final_prompt}
	\end{figure}
	
	In large language models (LLMs), prompt formulation defines how the forecasting task is communicated to the model and constrains the structure of the generated output. Since LLMs are primarily designed for natural language generation rather than numerical forecasting, the formulation of the instruction can substantially influence forecasting behavior. PRICE therefore employs a structured prompting strategy based on three components: \textit{Context}, \textit{Task}, and \textit{Format} (CTF).
	
	The \textit{Context} component specifies the forecasting setting by identifying the target asset as Bitcoin, the temporal granularity as hourly, and the data type as closing prices. It also provides the corresponding historical price sequence as the numerical input to the model. The \textit{Task} component defines the forecasting objective, namely predicting the next Bitcoin closing price from the provided context. The \textit{Format} component constrains the model output to a single numerical prediction without additional explanation or textual content. This structured formulation explicitly separates the forecasting context, objective, and expected output, thereby reducing ambiguity in the interaction between the task and the language model.
	
	To develop the forecasting instruction, multiple prompt variants are constructed for the same forecasting task. As illustrated in Fig.~\ref{prompt-engineering}, the process begins with a baseline instruction that specifies the forecasting objective and provides the relevant historical observations. To improve semantic clarity and task specification, a meta-instruction refinement procedure is subsequently applied. In this step, OpenAI's ChatGPT (GPT-4o) is used to refine the baseline instruction while preserving the original forecasting objective. 
	
	Building upon the refined instruction, additional prompt variants are developed using different prompting strategies, including Chain-of-Thought (CoT), Implicit Chain-of-Thought (iCoT), and few-shot prompting. All candidate prompts are evaluated under identical experimental conditions using the validation dataset described in Section~\ref{Prompt-Design}. The prompt formulation that achieves the strongest validation performance is selected as the CTF configuration shown in Fig.~\ref{final_prompt} and subsequently fixed for the remaining experiments. The resulting prompt is used consistently throughout the PRICE framework.
	
	\subsection{E: Exact Zero-Temperature Decoding}
	
	During inference, PRICE employs deterministic decoding with a temperature value of zero. This choice eliminates stochastic variation in the generation process and promotes consistent outputs across repeated inference steps. This property is particularly important for recursive multi-step forecasting, where each generated prediction is incorporated into the input for the subsequent step. Deterministic decoding therefore improves prediction stability and output reliability throughout the recursive forecasting process.
	
	
	\section{Experimental Setup}
	
	This section describes the datasets, implementation details, model configuration, and evaluation metrics used to assess the proposed LLM-based forecasting method.
	
	\subsection{Dataset and Preprocessing}
	
	\begin{table}[t]
		\centering
		\caption{Example format of the constructed BTC/USDT dataset.}
		\label{tab:data-format}
		\renewcommand{\arraystretch}{1.2}
		\begin{tabular}{cccccc}
			\hline
			Timestamp & Open & High & Low & Close & Volume\\
			\hline
			2024-06-01 00:00:00 & 67540.01 & 67585.89 & 67540.00 & 67573.41 & 11.67408\\
			2024-06-01 00:01:00 & 67573.41 & 67586.35 & 67561.40 & 67564.65 & 8.17416\\
			2024-06-01 00:02:00 & 67564.65 & 67582.90 & 67561.40 & 67582.90 & 9.24164\\
			\hline
		\end{tabular}
	\end{table}

	\begin{table}[t]
		\centering
		\caption{Chronological data split used for training, validation, and testing.}
		\label{tab:data-split}
		\renewcommand{\arraystretch}{1.4}
		\begin{tabular}{lccc}
			\hline
			Subset & Start Time & End Time & Ratio \\
			\hline
			Training & 2023-03-06 01:00 & 2024-08-01 16:00 & 80\% \\
			Validation & 2024-08-01 16:00 & 2024-10-05 00:00 & 10\% \\
			Test & 2024-10-05 00:00 & 2024-12-08 08:00 & 10\% \\
			\hline
		\end{tabular}
	\end{table}

	Historical cryptocurrency market data were collected from the Binance exchange using the CCXT library. We constructed a BTC/USDT forecasting dataset from raw minute-level OHLCV (Open, High, Low, Close, Volume) records collected between 2018 and 2024. Each observation contains timestamp information and the corresponding OHLCV attributes, as shown in Table~\ref{tab:data-format}. Bitcoin was selected due to its dominant role in cryptocurrency forecasting research, appearing in approximately 79\% of related studies~\cite{john2024cryptocurrency}.
	
	The collected data were preprocessed through timestamp normalization, duplicate removal, and missing value handling. Although the original dataset contains complete OHLCV information, this study focuses on univariate closing-price forecasting. The closing price was selected as the forecasting target because it represents the market value at the end of each interval and is commonly used in financial time-series forecasting. This setting allows the evaluation of LLMs based on historical price dependencies without introducing additional feature interactions.
	
	To reduce high-frequency market noise while preserving short-term market dynamics, minute-level observations were aggregated into hourly closing prices. This resolution provides a balance between temporal granularity and model efficiency by reducing short-term fluctuations while maintaining sufficient forecasting samples.
	
	Although raw data were collected from 2018 onward, the final experimental dataset uses observations from March 2023 to December 2024. This period was selected to provide a more consistent market environment and reduce potential distribution shifts caused by long-term changes in cryptocurrency market dynamics.
	
	The final dataset was divided chronologically into training, validation, and test sets using an 80\%/10\%/10\% ratio to preserve temporal ordering and prevent information leakage, as summarized in Table~\ref{tab:data-split}.
	
	\subsection{Model Architecture and Implementation Details}
	
	\begin{table}[t]
		\centering
		\caption{Fine-tuning hyperparameters}
		\label{tab:hyperparams}
		\renewcommand{\arraystretch}{1.3}
		\begin{tabular}{lc}
			\hline
			\textbf{Hyperparameter} & \textbf{Value} \\
			\hline
			Learning rate           & $5 \times 10^{-5}$ \\
			LR scheduler            & Cosine \\
			Warmup ratio            & 0.1 \\
			Training epochs         & 2 \\
			Batch size (per device) & 1 \\
			Gradient accumulation   & 4 \\
			Effective batch size    & 4 \\
			Max gradient norm       & 1.0 \\
			LoRA target             & Selected linear layers \\
			LoRA$+$ $\lambda$ ratio & 16.0 \\
			Training precision      & FP16 \\
			\hline
		\end{tabular}
	\end{table}

	Experiments were conducted using the LLaMA-3 8B-Instruct model~\cite{grattafiori2024llama}, implemented in its 4-bit quantized form\footnote{\url{https://huggingface.co/unsloth/llama-3-8b-Instruct-bnb-4bit}} to enable memory-efficient fine-tuning on limited hardware resources. Fine-tuning was performed using the \texttt{LLaMA-Factory} framework~\cite{zheng2024llamafactory} with LoRA (Low-Rank Adaptation)~\cite{hu2022lora}. Specifically, low-rank adapters were inserted into the selected linear layers while keeping the pretrained model parameters frozen, allowing efficient adaptation to cryptocurrency forecasting.
	
	Time series samples were generated using a sliding-window approach with stride 1. Each sample consists of 96 historical hourly closing-price observations used as input and the subsequent 8 observations as the forecasting target. Therefore, the input sequence length was set to \texttt{seq\_len}=96, corresponding to four days of historical data, while the forecasting horizon was set to \texttt{pred\_len}=8, corresponding to an 8-hour-ahead prediction task. This configuration targets short-term intraday cryptocurrency forecasting scenarios, where recent temporal dependencies are important due to the highly dynamic nature of digital asset markets.
	
	Training was conducted on a single NVIDIA GeForce RTX 2080 GPU with 11\,GB of VRAM. Unless otherwise specified, the default \texttt{LLaMA-Factory} training configuration was used. The main hyperparameters are summarized in Table~\ref{tab:hyperparams}.
	
	For comparative evaluation, transformer-based baselines were implemented using the official source code provided in the \texttt{Time-Series-Library (TSLib)} benchmark\footnote{\url{https://github.com/thuml/Time-Series-Library}}, following the recommended implementations to ensure fair and reproducible comparison under consistent experimental settings.
	
	\subsection{Evaluation Metrics}
	\label{sec:metrics}
	To quantitatively evaluate forecasting performance, multiple complementary metrics are employed. Let $\hat{y}_i$ and $y_i$ denote the predicted and ground-truth values at time step $i$, and let $\text{VNP}$ denote the number of valid numeric predictions.

	\subsubsection{Output Validity Assessment}
	
	Due to the generative nature of LLMs, generated outputs do not always conform to the required prediction format. Invalid outputs may contain non-numeric tokens (e.g., \texttt{"421.5abc"}), additional explanatory text (e.g., \texttt{"The predicted values are 421.5, 422.1"}), malformed numeric values, or a prediction sequence whose length differs from the required prediction length. Consequently, a post-processing step retains only outputs that consist exclusively of valid numeric values and match the expected prediction length. The number of retained predictions is referred to as the \emph{Valid Number of Predictions (VNP)}, and all evaluation metrics are computed exclusively on these valid predictions.
	
	\subsubsection{Scale-Dependent Metrics}
	
	The compared models differ fundamentally in their output scale: transformer-based models produce standardized outputs, whereas the LLM generates raw price values. To ensure fair comparison, scale-dependent metrics are computed after standardizing both predictions and targets using the mean and standard deviation estimated from the training set.
	
	\begin{equation}
		\text{MAE} =
		\frac{1}{\text{VNP}}
		\sum_{i=1}^{\text{VNP}}
		\lvert \hat{y}_i - y_i \rvert
		\label{eq:mae}
	\end{equation}
	
	\begin{equation}
		\text{MSE} =
		\frac{1}{\text{VNP}}
		\sum_{i=1}^{\text{VNP}}
		(\hat{y}_i - y_i)^2
		\label{eq:mse}
	\end{equation}
	
	\subsubsection{Scale-Independent Metrics}
	
	Scale-independent metrics are reported alongside scale-dependent metrics to provide complementary evaluation across models operating on different numerical ranges.
	
	\begin{equation}
		\text{MAPE} =
		\frac{100}{\text{VNP}}
		\sum_{i=1}^{\text{VNP}}
		\left|
		\frac{\hat{y}_i - y_i}{y_i}
		\right|
		\label{eq:mape}
	\end{equation}
	
	Note that SMAPE is bounded in $[0, 200]$ rather 
	than $[0, 100]$, since the denominator uses the 
	average of absolute values of both prediction 
	and ground truth. Values exceeding 100 indicate 
	predictions of opposite sign to the true values.
	
	\begin{equation}
		\text{SMAPE} =
		\frac{100}{\text{VNP}}
		\sum_{i=1}^{\text{VNP}}
		\frac{
			|\hat{y}_i - y_i|
		}{
			(|\hat{y}_i| + |y_i|)/2
		}
		\label{eq:smape}
	\end{equation}
	
	\begin{equation}
		\text{NRMSE} =
		\frac{
			\sqrt{
				\frac{1}{\text{VNP}}
				\sum_{i=1}^{\text{VNP}}
				(\hat{y}_i - y_i)^2
			}
		}{
			\frac{1}{\text{VNP}}
			\sum_{i=1}^{\text{VNP}}
			|y_i|
		}
		\label{eq:nrmse}
	\end{equation}
	
	
	\section{Experiments and Results}
	
	This section presents the experimental evaluation 
	of the proposed framework in two stages. First, 
	an ablation study on the validation set 
	investigates the impact of key design choices, 
	including input representation, forecasting 
	strategy, prompt design, and decoding temperature. 
	Subsequently, the optimized configuration is 
	compared with state-of-the-art transformer-based 
	and foundation forecasting models on both 
	validation and test sets.
	
	\subsection{Ablation Study}
	
	To systematically evaluate the contribution of 
	each design component, ablation experiments are 
	conducted by varying one factor at a time while 
	keeping all remaining settings fixed on the 
	validation set.
	
	\subsubsection{Effect of Input Representation}
	\label{Effect-of-Input-Representation}
	
	This experiment evaluates how different numerical representations of the input sequence affect forecasting performance under a zero-shot prompting setting. The objective is to determine whether simple transformations of raw price values improve the model's ability to generate accurate forecasts without task-specific fine-tuning.
	
	\paragraph{Experimental Setup:}

	\begin{figure}[ht]
		\centering
		\begin{promptbox}
			Based on the following 96 previous values, predict the next 8 values as a list.\\
			Return only the list of predicted numbers, without any explanation or extra text.\\
			-- input series --
		\end{promptbox}
		\caption{Fixed zero-shot prompt used to evaluate the effect of input sequence representation.}
		\label{prompt-input-sequence}
	\end{figure}
	
	To isolate the effect of input representation, a fixed minimal prompt is applied across all configurations (Fig.~\ref{prompt-input-sequence}), ensuring that performance differences are attributable solely to the numerical format of the input. Four representations are evaluated:
	
	\begin{itemize}
		
		\item \textbf{Raw Values:} Input prices with full decimal precision (e.g., \texttt{54504.06066666667}), preserving the original numerical form.
		
		\item \textbf{Rounded Values:} Input prices rounded to the nearest integer (e.g., \texttt{54504.06066666667} $\rightarrow$ \texttt{54504}), reducing numerical complexity and token fragmentation.
		
		\item \textbf{Standard Normalization:} Input values standardized using training-set statistics:
		\begin{equation}
			x' = \frac{x - \mu}{\sigma},
			\label{eq:standard_norm}
		\end{equation}
		where $\mu$ and $\sigma$ denote the training-set mean and standard deviation.
		
		\item \textbf{Instance Normalization:} Input values normalized independently for each sequence:
		\begin{equation}
			x'_i = \frac{x_i - \mu_i}{\sigma_i},
			\label{eq:instance_norm}
		\end{equation}
		where $\mu_i$ and $\sigma_i$ represent sequence-specific statistics for the $i$-th input.
		
	\end{itemize}

	Since the evaluated input representations operate on different numerical scales, scale-independent metrics are emphasized to enable fair comparison across configurations.

	\paragraph{Results and Discussion:}
	
	\begin{table*}[t]
		\centering
		\caption{Forecasting performance under different input sequence representations using MAPE, SMAPE, NRMSE, and VNP. Example inputs for each representation are shown in the last column.}
		\label{input_format_comparison}
		\renewcommand{\arraystretch}{1.3}
		\small
		\begin{tabular}{lccccc}
			\toprule
			\textbf{Format} & \textbf{MAPE} & \textbf{SMAPE} & \textbf{NRMSE} & \textbf{VNP} & \textbf{Example} \\
			\midrule
			Raw & 4.23 & 5.42 & 0.12 & 1441 & 54504.0606666667 \\
			Standard Normalization & 9.43 & 9.39 & 0.16 & 1441 & 0.8423 \\
			Instance Normalization & 548.36 & 132.19 & 1.41 & 1441 & -1.52752 \\
			\textbf{Rounded} & \textbf{3.01} & \textbf{3.05} & \textbf{0.05} & \textbf{1441} & 54504 \\
			\bottomrule
		\end{tabular}
	\end{table*}

	As shown in Table~\ref{input_format_comparison}, rounded values consistently achieve the lowest errors across all metrics, with MAPE of 3.01, SMAPE of 3.05, and NRMSE of 0.05, outperforming all other representations, including raw values.
	
	Standard normalization performs worse than both raw and rounded representations. A possible explanation is that it transforms naturally occurring price values into standardized numerical representations that are less consistent with the numerical distributions encountered during LLM pretraining. In contrast, rounded values preserve the original numerical scale while reducing unnecessary numerical precision, resulting in simpler numerical representations without losing magnitude information.
	
	Instance normalization exhibits the weakest performance across all metrics. The high MAPE value is partly attributable to the instability of percentage-based errors when normalized targets approach zero. However, the deterioration in SMAPE and NRMSE indicates that the degradation is not solely due to metric sensitivity. Unlike standard normalization, instance normalization rescales each input window independently, causing the same price value to be represented differently across samples. This inconsistency, together with the frequent occurrence of negative decimal values, may reduce numerical consistency across input windows and make learning stable forecasting patterns more difficult.
	
	Notably, VNP remains constant across all settings, indicating that the input representation primarily affects prediction accuracy rather than output validity. Based on these findings, the rounded integer representation is adopted for all subsequent experiments.

	\subsubsection{Effect of Learning Paradigm and Forecasting Strategy}
	
	This experiment investigates how learning paradigm and forecasting strategy influence LLM-based cryptocurrency time series forecasting performance.
	
	\paragraph{Experimental Setup:}
	We evaluate a $2\times2$ factorial design with two factors: learning paradigm and forecasting strategy. The learning paradigm factor includes two levels: zero-shot learning and fine-tuning, while the forecasting strategy factor includes two levels: single-step (direct) and multi-step (autoregressive) forecasting. Their combinations result in four experimental configurations, which are evaluated on the validation set.
	
	\begin{itemize}
		
		\item \textbf{Zero-Shot Paradigm:}  
		In the zero-shot setting, the LLM is directly used for forecasting without parameter updates. Predictions are generated solely based on the input prompt and the pre-trained knowledge of the model, providing a baseline for evaluating its inherent forecasting capability.
		
		\item \textbf{Fine-Tuning Paradigm:}  
		The pre-trained LLM is adapted to cryptocurrency time series using LoRA~\cite{hu2022lora}, enabling the model to learn domain-specific temporal patterns while keeping most of its parameters frozen.
		
	\end{itemize}
	
	Two forecasting strategies are then evaluated within each learning paradigm:
	
	\begin{itemize}
		
		\item \textbf{Single-Step (Direct) Forecasting:}
		The model receives 96 historical observations and directly generates the complete 8-step forecast in a single inference pass (Fig.~\ref{direct_prompt}).
		
		\begin{figure}[ht]
			\centering
			\begin{promptbox}
				Based on the following 96 previous values, predict the next 8 values as a list.\\
				Return only the list of predicted numbers, without any explanation or extra text.\\
				-- input series --
			\end{promptbox}
			\caption{Single-step forecasting prompt.}
			\label{direct_prompt}
		\end{figure}
		
		\item \textbf{Multi-Step (Autoregressive) Forecasting:}
		The model predicts one step at a time. After each prediction, the generated value is appended to the input sequence while the oldest observation is removed. This recursive process is repeated until the complete forecasting horizon is generated (Fig.~\ref{prompt_rolling}).
		
		\begin{figure}[ht]
			\centering
			\begin{promptbox}
				Based on the following 96 previous values, predict the next value.\\
				Return only the predicted number, without any explanation or extra text.\\
				-- input series --
			\end{promptbox}
			\caption{Multi-Step forecasting prompt.}
			\label{prompt_rolling}
		\end{figure}
		
	\end{itemize}
		
		\paragraph{Results and Discussion:}
		
		\begin{table}[t]
			\centering
			\caption{Comparison of forecasting performance across learning paradigms and forecasting strategies using MAE, MSE, and VNP.}
			\label{strategy_comparison}
			\renewcommand{\arraystretch}{1.3}
			\small
			\begin{tabular}{lccc}
				\toprule
				\textbf{Paradigm + Strategy} & \textbf{MAE} & \textbf{MSE} & \textbf{VNP} \\
				\midrule
				Zero-shot + Single-step   & 2063.62 & 7{,}405{,}313.89 & 1182 \\
				Zero-shot + Multi-step    & 1779.88 & 9{,}264{,}283.10 & 1441 \\
				Fine-tuned + Single-step  & 534.85  & 3{,}263{,}720.91 & 1440 \\
				\textbf{Fine-tuned + Multi-step}  & \textbf{445.60} & \textbf{423{,}562.43} & \textbf{1441} \\
				\bottomrule
			\end{tabular}
		\end{table}
		
		Table~\ref{strategy_comparison} summarizes forecasting performance across the four experimental configurations.
		
		\textbf{Effect of learning paradigm.}
		Fine-tuning substantially improves performance over zero-shot inference under both forecasting strategies. MAE decreases from 2063.62 to 534.85 under single-step forecasting (approximately 74\%) and from 1779.88 to 445.60 under multi-step forecasting (approximately 75\%), indicating that domain adaptation is important for capturing cryptocurrency market dynamics.
		
		\textbf{Effect of forecasting strategy.}
		Across both learning paradigms, the multi-step autoregressive strategy consistently achieves lower MAE than the single-step strategy. The advantage is particularly pronounced after fine-tuning, where multi-step forecasting substantially reduces MSE, suggesting more stable predictions than generating the entire forecasting horizon in a single inference step.
		
		In the zero-shot setting, however, multi-step forecasting produces lower MAE but higher MSE. This divergence suggests the presence of occasional extreme predictions, which disproportionately affect MSE due to its sensitivity to outliers. Nevertheless, MAE, being more robust to outliers, indicates that multi-step forecasting remains preferable on average.
		
		\textbf{Effect on output validity.}
		The VNP results further highlight the advantage of the multi-step strategy. Both multi-step configurations achieve perfect output validity (\texttt{VNP}=1441), whereas the zero-shot single-step configuration yields only 1182 valid predictions. This difference reflects the relationship between output length and generation reliability in LLMs: generating multiple future values simultaneously increases the likelihood of malformed or truncated outputs. By restricting each inference call to a single numeric prediction, the multi-step strategy improves structural output reliability.
		
		\textbf{Interaction between paradigm and strategy.}
		Interestingly, the multi-step strategy outperforms direct forecasting despite the well-known risk of recursive error accumulation. This suggests that, for short-horizon LLM forecasting, output reliability may represent a greater bottleneck than error propagation. While direct forecasting avoids recursive dependence, it requires longer structured generation, increasing the risk of invalid outputs. In contrast, the multi-step strategy trades limited error accumulation for substantially improved output validity and stability.
		
		Overall, the fine-tuned multi-step configuration achieves the strongest performance and is therefore adopted for subsequent experiments.
		
		\subsubsection{Effect of Prompt Design}
		\label{Prompt-Design}
		
		Prompt engineering refers to the systematic design and refinement of input instructions to guide Large Language Models (LLMs) toward more accurate and task-aligned responses \cite{marvin2023prompt}. This experiment examines how different prompt formulations influence LLM forecasting behavior in cryptocurrency price prediction.
		
		\paragraph{Experimental Setup:}
		As described in Section~\ref{Forecasting-Instruction}, multiple prompt configurations were developed through iterative reformulation, beginning with a minimal baseline instruction and progressively incorporating structured prompting strategies. All configurations were evaluated on the validation set under fixed experimental conditions. Most prompts were assessed using the fine-tuned multi-step configuration identified in earlier experiments. However, prompting strategies that are incompatible with fine-tuning, such as few-shot and chain-of-thought prompting, were evaluated under zero-shot inference; this distinction is explicitly reflected in the reported results.
		
		\begin{itemize}
			
			\item \textbf{Basic Prompt}
			
			\begin{figure}[ht]
				\centering
				\begin{promptbox}
					Based on the following 96 previous values, predict the next value.\\
					Return only the predicted number, without any explanation or extra text.\\
					-- input series --
				\end{promptbox}
				\caption{Basic prompt.}
				\label{prompt1}
			\end{figure}
			
			As shown in Figure~\ref{prompt1}, this configuration represents the basic prompt obtained through meta-instruction refinement. The instruction explicitly specifies the forecasting objective and constrains the response to a single numerical prediction without introducing additional prompt engineering strategies. Consequently, it serves as the reference configuration against which more advanced prompting formulations are evaluated.
			
			\item \textbf{CTF (Context--Task--Format) Prompt}
			
			\begin{figure}[ht]
				\centering
				\begin{promptbox}
					Using the following hourly Bitcoin closing prices, predict the next closing price based on comprehensive technical analysis.\\
					Return only one predicted number -- no explanation, no text.\\
					-- input series --
				\end{promptbox}
				\caption{CTF (Context--Task--Format) prompt.}
				\label{prompt2}
			\end{figure}

			To improve instruction clarity for numerical time series forecasting, a structured prompt formulation termed CTF (Context--Task--Format) was adopted based on established prompt engineering principles. Conventional role-based prompting---commonly structured around a role, task, and output format---typically assigns an analytical persona to guide model reasoning~\cite{white2023prompt}. However, in financial time series forecasting, selecting an appropriate role is less straightforward because the task simultaneously involves statistical pattern recognition, financial context, and numerical prediction. Rather than relying on role specification, the CTF formulation emphasizes explicit contextual grounding through three components: Context, which defines the target asset (Bitcoin), temporal granularity (hourly), and data type (closing prices); Task, which specifies the forecasting objective using an analytical framing (comprehensive technical analysis); and Format, which constrains the response to a single numerical prediction without explanatory text.
			
			Relative to the basic prompt, the CTF formulation provides clearer contextual specification of the forecasting task while preserving a simple instruction structure. The underlying assumption is that improved contextual grounding may enhance task alignment and forecasting consistency in numerical time series prediction.
			
			\item \textbf{Few-Shot Prompt}
			
			\begin{figure}[ht]
				\centering
				\begin{promptbox}
					Using the following hourly Bitcoin closing prices, perform forecasting based on historical patterns.\\
					Return only one predicted number -- no explanation, no text.\\
					\\
					Example 1:\\
					Input: "51718, ..., 62153"\\
					Output: "63162"\\
					\\
					Example 2:\\
					Input: "26065, ..., 27207"\\
					Output: "27303"\\
					\\
					Now predict based on this input:\\
					-- input series --
				\end{promptbox}
				\caption{Few-Shot prompt.}
				\label{prompt_fewshot}
			\end{figure}
			
			Few-shot prompting~\cite{logan2021cutting} conditions the model on a limited number of input--output demonstrations prior to inference. In the forecasting context, these demonstrations provide representative examples of how historical price sequences map to future values, thereby clarifying the expected prediction format and task structure.
			
			Unlike zero-shot prompting, which relies exclusively on natural-language instructions, few-shot prompting supplements task descriptions with concrete examples reflecting the desired input--output relationship. This additional contextual guidance may improve task alignment and prediction consistency by reducing ambiguity in how the forecasting objective is interpreted.
			
			For presentation clarity, Figure~\ref{prompt_fewshot} illustrates abbreviated examples, whereas all experiments were conducted using the complete 96-point input sequences.

			\item \textbf{Zero-Shot Chain-of-Thought Prompt}
			
			\begin{figure}[ht]
				\centering
				\begin{promptbox}
					Using the following hourly Bitcoin closing prices, analyze the data using technical analysis concepts (e.g., trends, momentum, and indicators).\\
					\\
					After your analysis, provide the next 8 predicted hourly closing prices. Clearly mark the output after the exact phrase:\\
					"final prediction:"\\
					\\
					Immediately after this phrase, output a Python list of exactly 8 numerical values (float), with no additional text or explanation.\\
					-- input series --
				\end{promptbox}
				\caption{Zero-Shot Chain-of-Thought prompt.}
				\label{prompt4}
			\end{figure}
			
			Zero-Shot Chain-of-Thought (CoT) prompting is a reasoning-oriented strategy that encourages models to generate intermediate reasoning steps without relying on task-specific demonstrations~\cite{wei2022chain}. Prior work by Kojima et al.~\cite{kojima2022large} showed that simple reasoning cues in natural-language prompts can improve performance in zero-shot settings by promoting more structured inference.
			
			As illustrated in Figure~\ref{prompt4}, the model is instructed to first analyze the temporal characteristics of the input series before generating a sequence of future predictions in a predefined output format. The strict formatting constraint facilitates automated extraction and evaluation while improving output consistency across samples. Since CoT prompting relies on inference-time reasoning rather than supervised adaptation, this configuration is evaluated under the zero-shot setting.
			
			\item \textbf{iCoT (Implicit Chain-of-Thought) Prompt}
			
			\begin{figure}[ht]
				\centering
				\begin{promptbox}
					Using the following hourly Bitcoin closing prices, predict the next closing price based on structured analysis of historical price dynamics.\
					\
					Your analysis should consider patterns commonly associated with technical analysis, including trend behavior, momentum signals, and support/resistance structures.\
					\
					Return only one predicted number --- no explanation, no text.\
					-- input series --
				\end{promptbox}
				\caption{iCoT (Implicit Chain-of-Thought) prompt.}
				\label{prompt3}
			\end{figure}
			
			Chain-of-Thought (CoT) prompting improves reasoning performance by encouraging models to generate intermediate reasoning steps prior to producing a final answer~\cite{wei2022chain}. However, in numerical forecasting tasks, explicit reasoning outputs may introduce practical limitations, including increased output variability and challenges for automated extraction of numerical predictions, particularly in fine-tuned forecasting pipelines where only the final forecast is required.
			
			Motivated by recent studies on implicit reasoning in LLMs~\cite{deng2024explicit,li2025implicit}, this study adopts an Implicit Chain-of-Thought (iCoT) prompting strategy tailored to financial time series forecasting. Rather than eliciting explicit reasoning traces, iCoT incorporates structured analytical guidance directly within the prompt while constraining the output to a single numerical prediction. The objective is to encourage reasoning-oriented processing without requiring externally visible reasoning steps.
			
			Compared with conventional CoT prompting, iCoT preserves a compact output format suitable for automated evaluation and fine-tuned numerical prediction. Furthermore, unlike training-based implicit reasoning approaches~\cite{deng2024explicit}, the proposed formulation requires no model modification and remains compatible with LoRA fine-tuning and autoregressive forecasting workflows.
			
		\end{itemize}
		
		\paragraph{Results and Discussion:}
		
		\begin{table*}[t]
			\centering
			\caption{Forecasting performance comparison across prompt engineering strategies under fine-tuned and zero-shot inference settings. Lower MAE and MSE indicate better performance, whereas higher VNP indicates greater output validity. Best results within each configuration group are highlighted in bold.}
			\label{prompt_comparison}
			\renewcommand{\arraystretch}{1.25}
			\small
			\begin{tabular}{llcccc}
				\toprule
				\textbf{Prompt Strategy} & 
				\textbf{Paradigm} & 
				\textbf{Model} & 
				\textbf{MAE $\downarrow$} & 
				\textbf{MSE $\downarrow$} & 
				\textbf{VNP $\uparrow$} \\
				\midrule
				
				\multicolumn{6}{l}{\textit{Fine-tuned configurations}} \\
				Basic & Fine-tuned & LLaMA & 
				445.60 & 423{,}562.43 & 1441 \\
				
				iCoT & Fine-tuned & LLaMA & 
				436.68 & 412{,}852.85 & 1441 \\
				
				CTF & Fine-tuned & LLaMA & 
				\textbf{432.83} & \textbf{402{,}655.83} & 1441 \\
				
				\midrule
				
				\multicolumn{6}{l}{\textit{Zero-shot configurations}} \\
				CTF & Zero-shot & GPT-mini & 
				\textbf{519.12} & \textbf{562{,}296.72} & 1441 \\
				
				CoT & Zero-shot & GPT-mini & 
				537.81 & 3{,}425{,}467.31 & 1441 \\
				
				CoT & Zero-shot & GPT-nano & 
				958.17 & 2{,}243{,}561.01 & 1434 \\
				
				Few-Shot & Zero-shot & LLaMA & 
				3{,}296.07 & $1.72 \times 10^{7}$ & 1441 \\
				
				CoT & Zero-shot & LLaMA & 
				13{,}559.24 & $7.06 \times 10^{8}$ & 1403 \\
				
				\bottomrule
			\end{tabular}
		\end{table*}
		
		Table~\ref{prompt_comparison} summarizes forecasting performance across all evaluated prompt configurations on the validation set. The results indicate that prompt formulation substantially influences forecasting accuracy, with effects varying across inference paradigms, model capacities, and the degree of alignment between prompt structure and the numerical characteristics of the forecasting task.
		
		Among the fine-tuned configurations sharing the same LLaMA backbone, the CTF prompt achieves the lowest MAE (432.83) and MSE (402,655.83), followed by iCoT (MAE 436.68) and the basic prompt (MAE 445.60). Although instruction design might be expected to become less influential after supervised adaptation, this ranking demonstrates that prompt engineering continues to contribute measurably to forecasting performance even under fine-tuning, where the same instruction is presented during both training and inference. The observed improvement suggests that clearer contextual specification and structured task framing enhance alignment with the forecasting objective despite prior model adaptation. The iCoT formulation also outperforms the baseline prompt but remains slightly inferior to CTF, indicating that concise contextual grounding may be more effective than elaborate reasoning-oriented guidance in the fine-tuned setting. Notably, VNP remains constant at 1441 across all fine-tuned configurations, suggesting that prompt complexity does not materially affect output validity after fine-tuning.
		
		Under zero-shot inference, the CTF prompt with GPT-mini achieves the strongest performance among zero-shot configurations (MAE 519.12), indicating that carefully structured instructions can partially mitigate the absence of task-specific fine-tuning. Nevertheless, a clear performance gap remains between fine-tuned and zero-shot settings, reinforcing the dominant role of supervised adaptation in forecasting accuracy. GPT-nano exhibits weaker performance, suggesting that smaller-capacity models are less effective at leveraging structured prompting strategies in high-volatility financial forecasting.
		
		The few-shot configuration produces the largest prediction errors among zero-shot methods despite maintaining perfect output validity. This finding is notable given the widespread effectiveness of few-shot prompting in language tasks. In volatile financial time series, however, a limited number of demonstrations may inadequately represent evolving market regimes, potentially introducing anchoring effects that bias predictions toward demonstrated historical patterns rather than improving generalization.
		
		CoT prompting is evaluated across three model variants to examine the interaction between reasoning-oriented prompting and model capacity. LLaMA under zero-shot CoT produces the weakest performance across all configurations and the lowest VNP (1,403), suggesting that extended reasoning-style prompting may reduce numerical consistency and output reliability in models lacking robust zero-shot reasoning capability. GPT-nano and GPT-mini perform comparatively better, indicating that greater model capacity partially mitigates these limitations, although both remain inferior to the fine-tuned CTF configuration.
		
		Overall, the findings suggest that structured and concise instruction design is more effective than reasoning-intensive or example-based prompting for cryptocurrency time series forecasting. More broadly, prompt engineering strategies successful in general language tasks do not necessarily transfer effectively to numerical forecasting problems and should therefore be evaluated in a task-specific manner. Importantly, the results further demonstrate that prompt engineering remains consequential even after fine-tuning, challenging the assumption that instruction design becomes negligible once model adaptation has been performed. Based on these findings, the CTF formulation is adopted as the final instruction design for all subsequent experiments.

		\subsubsection{Effect of Temperature}
		
		The temperature parameter controls the stochasticity of autoregressive generation by modulating the probability distribution over candidate tokens. Lower temperature values favor high-probability token selection, resulting in more deterministic outputs, whereas higher temperatures increase output diversity at the expense of greater uncertainty and reduced prediction stability \cite{minaee2024large}.
		
		\paragraph{Experimental Setup:}
		
		In this experiment, temperature sensitivity was evaluated under the fine-tuning paradigm using the multi-step forecasting strategy and the prompt configuration identified as optimal in previous experiments. Temperature values were varied from 0 to 1.9 in fixed increments, and forecasting performance was assessed on the validation set using standard error-based evaluation metrics.
		
		The default temperature setting in the \texttt{LLaMA Factory} framework is 0.95, which was consistently adopted in prior experiments. Consequently, this experiment aims to examine the extent to which deviations from the default sampling behavior influence numerical forecasting performance under otherwise identical model and prompting conditions.
		
		\paragraph{Results and Discussion:}
		
		\begin{table}[t]
			\centering
			\caption{Effect of temperature on forecasting performance under the fine-tuned multi-step forecasting setting using the CTF prompt. Lower MAE and MSE indicate better performance, whereas higher VNP indicates greater output validity. Best results are highlighted in bold.}
			\label{temperature_experiment}
			\renewcommand{\arraystretch}{1.25}
			\small
			\begin{tabular}{cccc}
				\toprule
				\textbf{Temperature} & 
				\textbf{MAE $\downarrow$} & 
				\textbf{MSE $\downarrow$} & 
				\textbf{VNP $\uparrow$} \\
				\midrule
				0.0  & \textbf{415.89} & \textbf{385,195.52} & 1441 \\
				0.2  & 417.32 & 386,571.30 & 1441 \\
				0.6  & 422.83 & 391,328.30 & 1441 \\
				0.8  & 426.77 & 397,966.41 & 1441 \\
				0.95 & 432.48 & 404,533.49 & 1441 \\
				1.0  & 434.25 & 405,248.15 & 1441 \\
				1.9  & 611.14 & 743,941.20 & 1441 \\
				\bottomrule
			\end{tabular}
		\end{table}
		
		Table~\ref{temperature_experiment} summarizes the effect of temperature on forecasting performance under the fine-tuned multi-step setting. The results indicate a clear relationship between sampling stochasticity and forecasting accuracy, with lower temperature values consistently improving predictive performance.
		
		The best performance is achieved at temperature 0.0, yielding the lowest MAE and MSE. As temperature increases, performance degrades in a near-monotonic manner. Relative to the default setting used in prior experiments (0.95), deterministic decoding reduces MAE by approximately 3.8\% and MSE by 4.8\%, indicating that lower stochasticity improves numerical forecasting accuracy.
		
		This behavior is consistent with the deterministic nature of time series forecasting, where accurate continuation of temporal structure is required. In contrast to open-ended text generation, higher temperature values introduce sampling variability that can disrupt learned numerical regularities, leading to less stable predictions. The sharp degradation at temperature 1.9 further confirms the sensitivity of forecasting performance to excessive randomness.
		
		Notably, VNP remains constant across all settings, indicating that temperature affects prediction accuracy rather than output validity. Overall, the results suggest that deterministic or near-deterministic decoding is preferable for LLM-based cryptocurrency forecasting. Based on these findings, temperature 0.0 is adopted for subsequent experiments.
		
		\subsection{Comparative Evaluation}

		\begin{table}[t]
			\centering
			\caption{Performance comparison between PRICE and the evaluated transformer-based and time-series foundation models for multi-step forecasting ($\mathrm{pred\_len}=8$) on the validation and test sets. Lower values indicate better performance. Best results are highlighted in bold.}
			\label{alg_comparison_all_pred8}
			
			\renewcommand{\arraystretch}{1.15}
			\resizebox{\textwidth}{!}{%
				\begin{tabular}{lcccccccccc}
					\toprule
					\multirow{2}{*}{\textbf{Method}}
					& \multicolumn{5}{c}{\textbf{Validation Set}}
					& \multicolumn{5}{c}{\textbf{Test Set}} \\
					
					\cmidrule(lr){2-6}
					\cmidrule(lr){7-11}
					
					& \textbf{MAE}
					& \textbf{MSE}
					& \textbf{MAPE}
					& \textbf{SMAPE}
					& \textbf{NRMSE}
					& \textbf{MAE}
					& \textbf{MSE}
					& \textbf{MAPE}
					& \textbf{SMAPE}
					& \textbf{NRMSE} \\
					
					\midrule
					
					Transformer
					& 0.05 & 0.0045 & 3.57 & 3.51 & 0.04
					& 0.29 & 0.17 & 9.97 & 10.90 & 0.18 \\
					
					Autoformer
					& 0.09 & 0.01 & 6.08 & 6.08 & 0.08
					& 0.10 & 0.0178 & 4.78 & 4.85 & 0.06 \\
					
					Crossformer
					& 0.12 & 0.02 & 8.57 & 8.07 & 0.10
					& 0.65 & 0.78 & 22.74 & 27.60 & 0.38 \\
					
					Pyraformer
					& 0.04 & 0.0030 & 2.77 & 2.74 & 0.04
					& 0.35 & 0.23 & 11.71 & 12.99 & 0.21 \\
					
					PatchTST
					& 0.03 & 0.0022 & 2.23 & 2.23 & 0.03
					& 0.04 & 0.0035 & 1.78 & 1.79 & 0.03 \\
					
					TimesNet
					& 0.04 & 0.0032 & 2.79 & 2.83 & 0.04
					& 0.04 & 0.0038 & 1.82 & 1.83 & 0.03 \\
					
					Lag-Llama
					& 0.11 & 0.02 & 9.89 & 9.76 & 0.12
					& 0.16 & 0.05 & 7.09 & 7.38 & 0.09 \\
					
					Chronos
					& 0.03 & 0.0017 & 2.52 & 2.52 & 0.03
					& 0.04 & 0.0040 & 1.77 & 1.78 & 0.03 \\
					
					DeepSeek
					& 0.10 & 0.0189 & 9.71 & 9.39 & 0.13
					& 0.13 & 0.0344 & 5.77 & 5.97 & 0.08 \\
					
					\textbf{PRICE}
					& \textbf{0.02}
					& \textbf{0.0014}
					& \textbf{2.19}
					& \textbf{2.19}
					& \textbf{0.03}
					& \textbf{0.03}
					& \textbf{0.0029}
					& \textbf{1.53}
					& \textbf{1.53}
					& \textbf{0.02} \\
					
					\bottomrule
				\end{tabular}%
			}
		\end{table}
		
		Using the optimal configuration identified through the ablation study, PRICE is compared with transformer-based baselines and time-series foundation models under identical experimental conditions. Table~\ref{alg_comparison_all_pred8} reports the forecasting performance on both the validation and test sets.
		
		On the validation set, PRICE achieves the lowest error for MAE, MSE, MAPE, and SMAPE, while tying with PatchTST and Chronos for the lowest NRMSE. Specifically, PRICE obtains an MAE of 0.02, MSE of 0.0014, MAPE of 2.19, SMAPE of 2.19, and NRMSE of 0.03. Among the evaluated baselines, PatchTST and Chronos provide the strongest overall performance, both achieving an MAE of 0.03. Nevertheless, PRICE maintains lower errors across the other reported metrics, indicating stronger predictive precision under the selected multi-step forecasting setting.
		
		On the test set, PRICE achieves the lowest values across all five evaluation metrics, with an MAE of 0.03, MSE of 0.0029, MAPE of 1.53, SMAPE of 1.53, and NRMSE of 0.02. PatchTST and TimesNet constitute the strongest competing baselines, both achieving an MAE of 0.04, with SMAPE values of 1.79 and 1.83, respectively. The consistently lower errors achieved by PRICE indicate strong generalization to the held-out test period. Moreover, the relatively stable performance of PRICE across the validation and test sets contrasts with the substantial degradation observed for several baseline models.
		
		Several baseline models exhibit substantial performance degradation between the validation and test sets. Crossformer, for example, increases from an SMAPE of 8.07 to 27.60, while Pyraformer increases from 2.74 to 12.99 and the vanilla Transformer from 3.51 to 10.90. These changes indicate greater sensitivity to temporal distributional differences between the evaluation periods. In contrast, PRICE maintains comparatively stable performance across the two evaluation sets.
		
		Among the evaluated time-series foundation models, Chronos achieves competitive performance, particularly on the validation set, whereas Lag-LLaMA exhibits substantially higher forecasting errors across both evaluation sets. Notably, PRICE is built upon LLaMA-3 8B, a general-purpose large language model originally pretrained primarily on text rather than on time-series data, whereas Chronos and Lag-LLaMA are specifically designed or pretrained for time-series forecasting. Despite this difference in pretraining objectives, PRICE achieves competitive or superior performance relative to these specialized time-series foundation models, obtaining the lowest forecasting errors on the test set.
		
		DeepSeek-V3.2 provides an additional reference point for evaluating the capability of a general-purpose LLM for numerical forecasting. The configuration used for DeepSeek-V3.2 was informed, where applicable, by the findings of the ablation study, incorporating components that demonstrated beneficial effects in the PRICE experiments. Its relatively competitive performance suggests that a general-purpose LLM can capture meaningful patterns from numerical forecasting inputs; however, the substantial performance gap between DeepSeek-V3.2 and PRICE highlights the potential benefits of systematic adaptation, particularly parameter-efficient fine-tuning, for this forecasting task. These results further suggest that effective domain-specific adaptation can improve the forecasting capability of a general-purpose language model, enabling it to serve as a competitive forecaster for short-term Bitcoin price forecasting.
		
		Overall, PRICE achieves the lowest or tied-lowest errors across the reported validation metrics and the lowest errors across all test metrics. These results demonstrate that systematically combining parameter-efficient fine-tuning, recursive inference, integer-rounded numerical representation, structured prompting, and deterministic decoding enables an adapted general-purpose LLM to outperform the evaluated transformer-based and time-series foundation models for short-term Bitcoin price forecasting.

	\section{Conclusion}
	
	This study presented PRICE, a systematic framework for adapting large language models to short-term Bitcoin price forecasting. Built on a 4-bit quantized LLaMA-3 8B model, PRICE integrates five complementary adaptation components: parameter-efficient fine-tuning with LoRA, recursive multi-step inference, integer-rounded numerical representation, Context-Task-Format (CTF) prompting, and exact zero-temperature decoding. Through controlled ablation experiments, we systematically examined the contribution of these adaptation choices to forecasting accuracy and prediction reliability.
	
	The experimental results demonstrate that each adaptation dimension can materially influence LLM-based numerical forecasting. LoRA fine-tuning substantially improves predictive accuracy relative to zero-shot inference, while recursive multi-step inference achieves lower forecasting errors and higher output validity than direct multi-step generation. Integer-rounded numerical representation consistently outperforms raw, standard-normalized, and instance-normalized representations, suggesting that preserving the original price scale while reducing unnecessary numerical precision provides a more effective representation for LLM-based financial forecasting. Furthermore, CTF prompting achieves the strongest performance among the evaluated prompting strategies in the fine-tuned setting, demonstrating that prompt formulation remains consequential even after supervised adaptation. Exact zero-temperature decoding further improves prediction stability during recursive forecasting by eliminating sampling-induced variability.
	
	Comparative evaluation against transformer-based forecasting architectures and time-series foundation models demonstrates the effectiveness of PRICE under the evaluated conditions. PRICE achieves the lowest or tied-lowest errors across the reported validation metrics and the lowest errors across all test metrics. Notably, PRICE is based on LLaMA-3 8B, a general-purpose language model originally pretrained primarily on text rather than specifically on time-series data, yet it achieves competitive or superior performance relative to specialized time-series foundation models. Moreover, PRICE maintains comparatively stable performance across the validation and test periods, whereas several baseline models exhibit substantial degradation, indicating greater sensitivity to temporal distributional differences.
	
	Overall, the findings demonstrate that the effectiveness of an LLM for numerical forecasting depends not only on the underlying model but also on how the model is adapted, how numerical data are represented, how the forecasting task is formulated, and how predictions are generated. The results therefore support the central premise of PRICE: adaptation choices that are often treated as implementation details can represent important determinants of forecasting accuracy, output reliability, and robustness. Systematic evaluation of these choices can enable general-purpose LLMs to serve as competitive forecasters even when they are not originally designed or pretrained specifically for time-series prediction.
	
	This study is subject to several limitations. The experiments focus on a single asset (Bitcoin), a fixed temporal granularity (hourly data), a fixed forecasting horizon, and one LLM backbone (LLaMA-3 8B). In addition, the reported findings are based on the evaluated market periods and model configurations and therefore should not be interpreted as universally applicable to all financial assets or forecasting horizons. Future research could investigate the generalizability of the PRICE framework across multiple assets, temporal resolutions, forecasting horizons, and LLM backbones. Extending the analysis to additional financial and non-financial time series could further determine whether the observed benefits of numerical representation, prompt formulation, recursive inference, and deterministic decoding extend beyond Bitcoin forecasting.

	\FloatBarrier
	\bibliographystyle{unsrt}
	\bibliography{btc_refs}
	
\end{document}